\documentclass[
twocolumn,
]{ceurart}

\usepackage{listings}
\usepackage{url}

\begin{document}

\copyrightyear{2022}
\copyrightclause{Copyright for this paper by its authors.
  Use permitted under Creative Commons License Attribution 4.0
  International (CC BY 4.0).}

\conference{EBeM'22: AI Evaluation Beyond Metrics, 
  July 24, 2022, Vienna, Austria}

 \title{Evaluating Understanding on Conceptual Abstraction Benchmarks}

 \author[1]{Victor Vikram Odouard}[
 email=vo47@cornell.edu,
 ]
\author[1]{Melanie Mitchell}[
 email=mm@santafe.edu,
 ]
 \address[1]{Santa Fe Institute, 1399 Hyde Park Road, Santa Fe, NM 87501 USA}

\begin{abstract}
A long-held objective in AI is to build systems that understand concepts in a humanlike way. Setting aside the difficulty of building such a system, even trying to evaluate one is a challenge, due to present-day AI’s relative opacity and its proclivity for finding shortcut solutions. This is exacerbated by humans’ tendency to anthropomorphize, assuming that a system that can recognize one instance of a concept must also understand other instances, as a human would. In this paper, we argue that understanding a concept requires the ability to use it in varied contexts. Accordingly, we propose systematic evaluations centered around concepts, by probing a system’s ability to use a given concept in many different instantiations. We present case studies of such an evaluations on two domains---RAVEN (inspired by Raven's Progressive Matrices) and the Abstraction and Reasoning Corpus (ARC)---that have been used to develop and assess abstraction abilities in AI systems.  Our \textit{concept-based} approach to evaluation reveals information about AI systems that conventional test sets would have left hidden.
\end{abstract}


\maketitle

\section{Introduction}

What unites chain-link fences, high prices, entrance exams, and import tariffs? They are all different kinds of \emph{barriers}. Your understanding of physical barriers may have helped you quickly intuit how chess pieces move (and the fundamental difference between the knight and the other pieces) from very few examples. It may have helped you relate to a friend struggling with credit card debt, even when your obstacles are very different. It may have helped you describe how being jet-lagged sometimes feels like ``hitting a wall.''  These examples illustrate the importance of abstract concepts in few-shot learning, generalization, emotional intelligence, and communication.  Such examples display the intuition behind Barsalou's definition of a concept: ``a competence or disposition for generating infinite conceptualizations of a category'' \cite{barsalou2020challenges}.  In short, understanding the world entails being able to recognize and generate concepts in both concrete and abstract forms.  

Early pioneers suggested that their AI summer project might lead to blueprints for machines that could ``form abstractions and concepts'' \cite{mccarthy2006proposal}. More than six decades later, AI systems are still extremely limited in this regard: they have yet to surmount the ``barrier'' of understanding \cite{mitchell2019artificial}.

Evaluating a system's understanding of concepts and abstractions is challenging. AI systems are known to be susceptible to shortcut learning, such as recognizing pictures of animals by looking for blurry backgrounds \cite{landecker2013interpeting} or pictures of cows by looking at surrounding landscapes \cite{geirhos2020shortcut}. More insidiously, certain image classifiers can be fooled into classifying, say, school buses as ostriches by changing the picture in ways indiscernible to human viewers \cite{szegedy2013intriguing}.

In this paper, we propose systematic assessments centered around concepts---a \textit{concept-based} approach---to evaluate understanding in AI systems.  This approach involves (1) identifying a set of concepts a system should know and (2) designing sets of questions probing for the grasp of these concepts using a variety of instantiations of each concept.   

One of the important pillars of the traditional train/test paradigm in machine learning---that the training and test sets be independent and identically distributed (IID)---is violated with our concept-based evaluation method. In order to probe understanding by creating varied concept instantiations, the examples used for evaluation may not be drawn from the same ``distribution'' as the training set. Furthermore, the examples in evaluation set will likely not be independent in any sense, since they are created by varying specific concepts.  In two case studies, we find that our evaluation method reveals important information about a system's ability to understand concepts that might be hidden using a conventional IID test set. 

\begin{figure*}[t]
\begin{centering}
    \includegraphics[width=.55\textwidth]{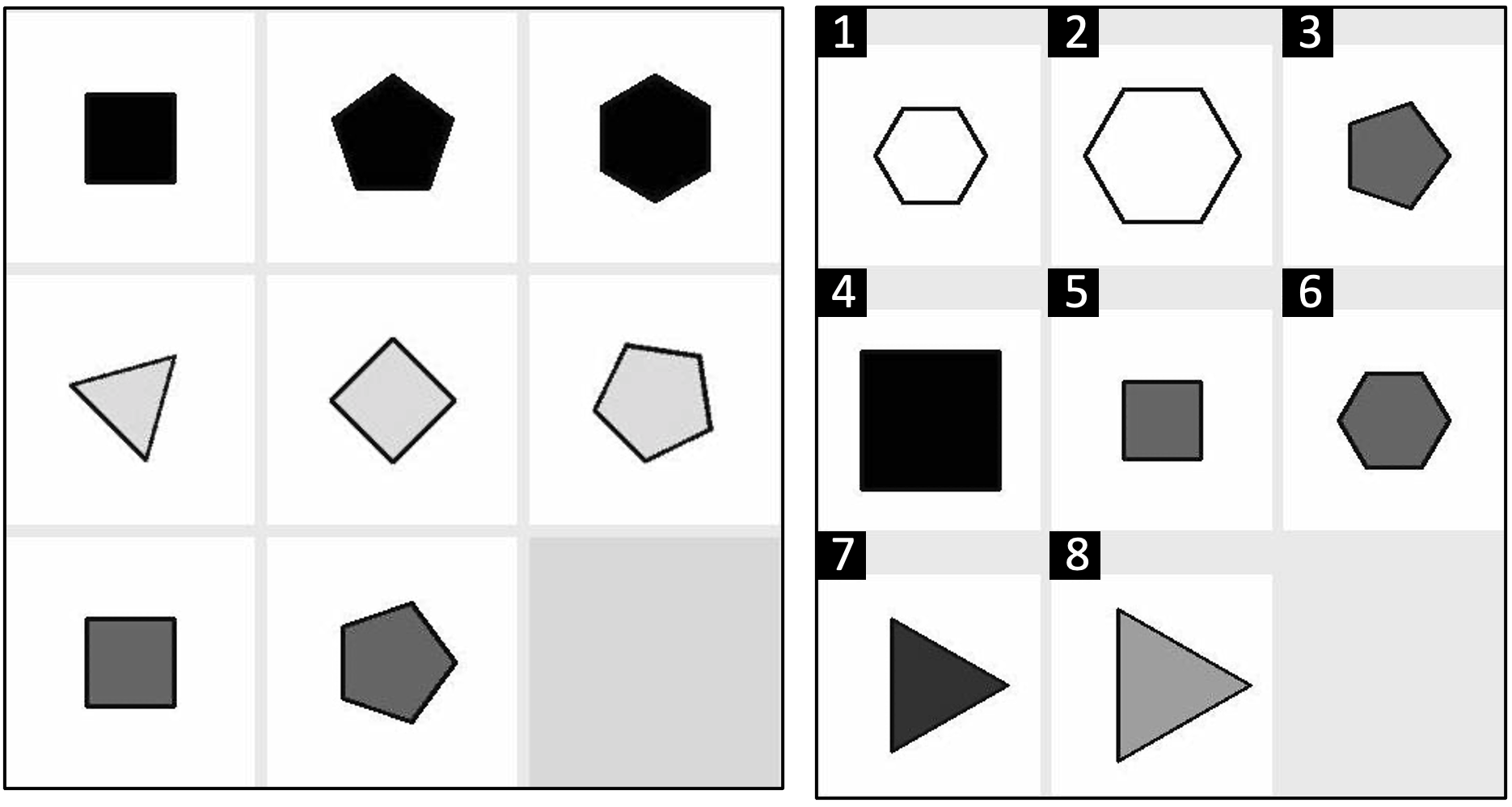} 
    \caption{A sample problem in the RAVEN domain. Each row gives polygons with increasing number of sides, with size and color (i.e., gray scale) staying fixed; the correct answer is choice 6.} 
\end{centering}    
    \label{RAVEN1}
\end{figure*}

\begin{figure*}[t]
\begin{centering}
    \includegraphics[width=.55\textwidth]{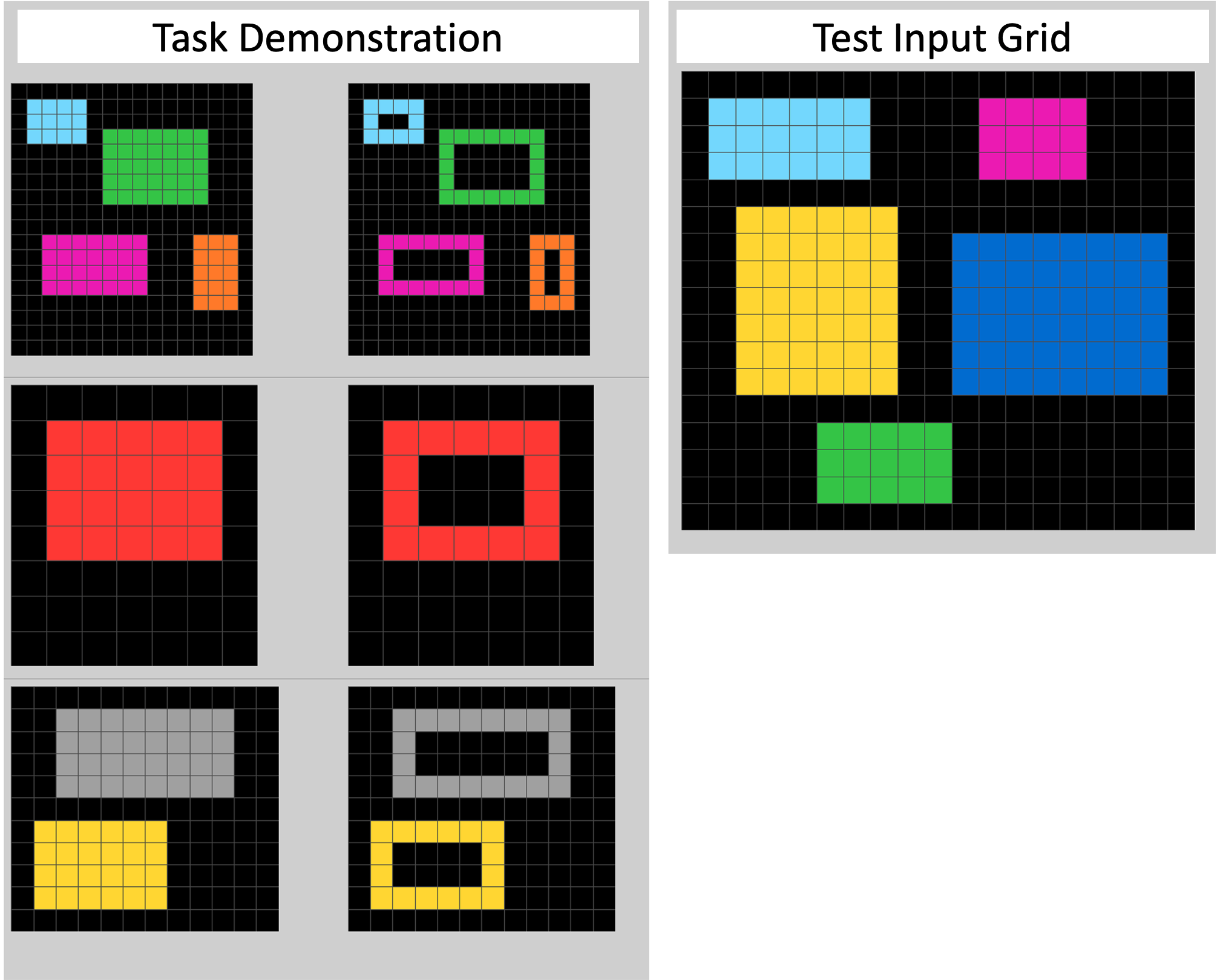} \caption{A sample problem (``task'') in the ARC domain. The solver's challenge is to generate a grid that transforms the test input grid in the same way as in the task demonstrations.  Best viewed in color.} \label{ARC1}
\end{centering}
\end{figure*}

We created concept-based evaluations for two domains that have been used to develop and assess conceptual abstraction abilities in AI systems: RAVEN \cite{zhang2019raven} (inspired by Raven's Progressive Matrices (RPMs) \cite{raven1938ravens}) and the Abstraction and Reasoning Corpus (ARC) \cite{chollet2019intelligence}. Figure~\ref{RAVEN1} shows a sample problem in the RAVEN domain.  Each such problem consists of a three-by-three matrix (Figure~\ref{RAVEN1} left) in which each of 8 matrix components is a figure involving geometric shapes, with some relationship between the figures in the rows and columns.  The ninth component is missing, and the task is to fill in the missing component with one of a set of eight candidate answers (Figure~\ref{RAVEN1} right).  

ARC problems (termed ``tasks'' in \cite{chollet2019intelligence}) present a number of ``demonstration'' pairs of grids which are related via a transformation rule, asking the solver to ``do the same thing'' (i.e., apply the same transformation) to a new ``test'' input grid. Figure~\ref{ARC1} shows a sample task in the ARC domain. The solver's challenge is to generate a new grid that transforms the test input grid analogously to the transformations in the  demonstration grids.  The concepts used in the ARC domain were inspired by Spelke's proposals for core knowledge systems \cite{spelke2007core} such as spatio-temporal relations (inside, above, next-to), object attributes (shape, size, color, boundary), transformations (rotate, shift, extend), and more general relations (progression, sameness, part-whole).  Notably, ARC tasks require the solver to \textit{generate} an answer, rather than choose among given candidate answers, as in RAVEN, providing the potential for more insight into the understanding of the solver \cite{chollet2019intelligence}.

\section{Prior Results on RAVEN}

The RAVEN domain was inspired by Raven's Progressive Matrices (RPMs), a kind of IQ test that has been used to measure ``fluid intelligence'' in humans for many decades \cite{raven1938ravens}.  There have been numerous efforts to apply AI and machine learning methods to RPM-like problems (e.g., \cite{zhang2019raven,barrett2018measuring,benny2021scale,hu2021stratified,lovett2017modeling,spratley2020closer,wang2015automatic,wu2020scattering}, among many others).  Recently many groups have applied deep neural networks (DNNs) to such problems, but given that DNNs need large numbers of training examples, these efforts require methods for procedural generation of these examples.  The creators of the RAVEN dataset \cite{zhang2019raven} developed one such method (another method was used to generate the PGM dataset \cite{barrett2018measuring}).  To generate a RAVEN problem, the system sampled from a hierarchical stochastic image grammar \cite{zhang2019raven}, which offered different possible layouts for the matrix components (e.g., center, inside/outside, grid), and within each layout it offered a choice of shapes (e.g., circle, square, triangle, pentagon) with different attributes to be chosen (e.g., color, size, angle), where each attribute is constrained to be one of a small number of values.  The grammar also enforced one of a choice of relationships between matrix elements in a row (e.g., constant, progression, arithmetic); see \cite{zhang2019raven} for details. The authors generated 70,000 problems total, splitting RAVEN into 42,000 training, 14,000 validation, and 14,000 test examples.

In the paper detailing the RAVEN dataset, Zhang et al. \cite{zhang2019raven} reported human performance on RAVEN's test set at 84\% accuracy on average.  Several subsequent papers reported deep learning methods which surpassed human performance on this dataset (e.g., \cite{zhang2019learning,zhuo2020solving}).

The original RAVEN dataset, however, had a bias in its answer-generation method: answer choices were generated by taking the correct answer and modifying an attribute, allowing solvers to take the majority vote for each attribute to get the correct answer. In fact, networks trained \emph{solely} on the answer choices could attain over 90\% accuracy \cite{hu2021stratified}. To remedy this shortcoming, other groups generated modified versions of the answer choices in RAVEN using methods that that seem to be less exploitable. The new versions of RAVEN included RAVEN-FAIR \cite{benny2021scale} and I-RAVEN \cite{hu2021stratified}. Several groups have since reported test-set accuracies on these new versions that significantly surpass the human performance benchmark of 84\% (e.g., \cite{benny2021scale,wu2020scattering,malkinski2020multi}).

\section{Concept-Based Evaluations for RAVEN}

When a program (e.g., a DNN) exhibits high accuracy on the RAVEN dataset, does the program \textit{understand} the concepts expressed in the problems it solved, as a human would? And when a program for solving ARC problems correctly solves a task, to what extent is the program capturing the abstract reasoning abilities the dataset's name implies?

As we have argued above, the way to answer these questions is to evaluate these programs on systematic variations of the \textit{concepts} that they purport to understand.  Neither the RAVEN nor ARC datasets (nor any other abstraction datasets that we are aware of) provides this kind of evaluation.  In this section we demonstrate how such an evaluation can be carried out on programs that score high on the RAVEN test set. 

\begin{figure*}[h]
\begin{centering}
    \includegraphics[width=.55\textwidth]{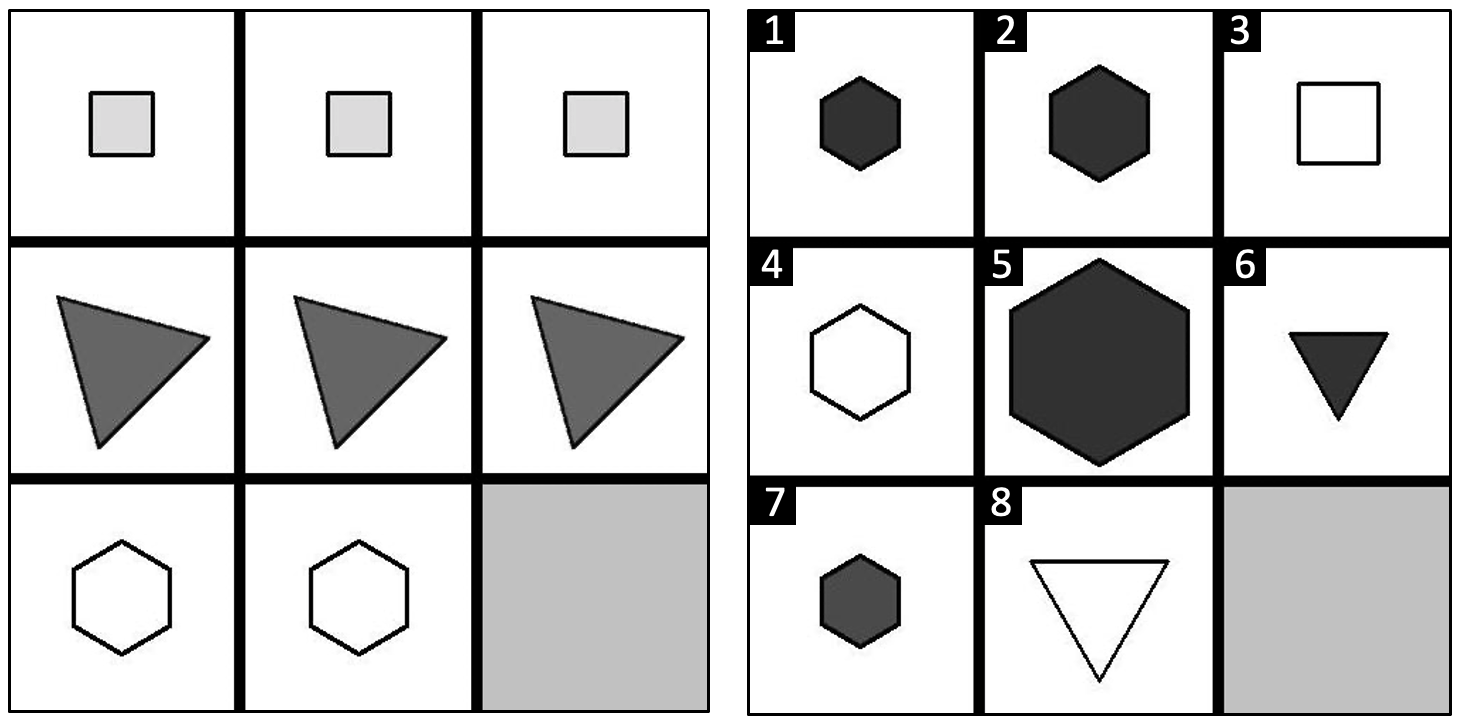} 
    
    \Large \textbf{(a)}
    
    \vspace*{.1in}
    
    \includegraphics[width=.55\textwidth]{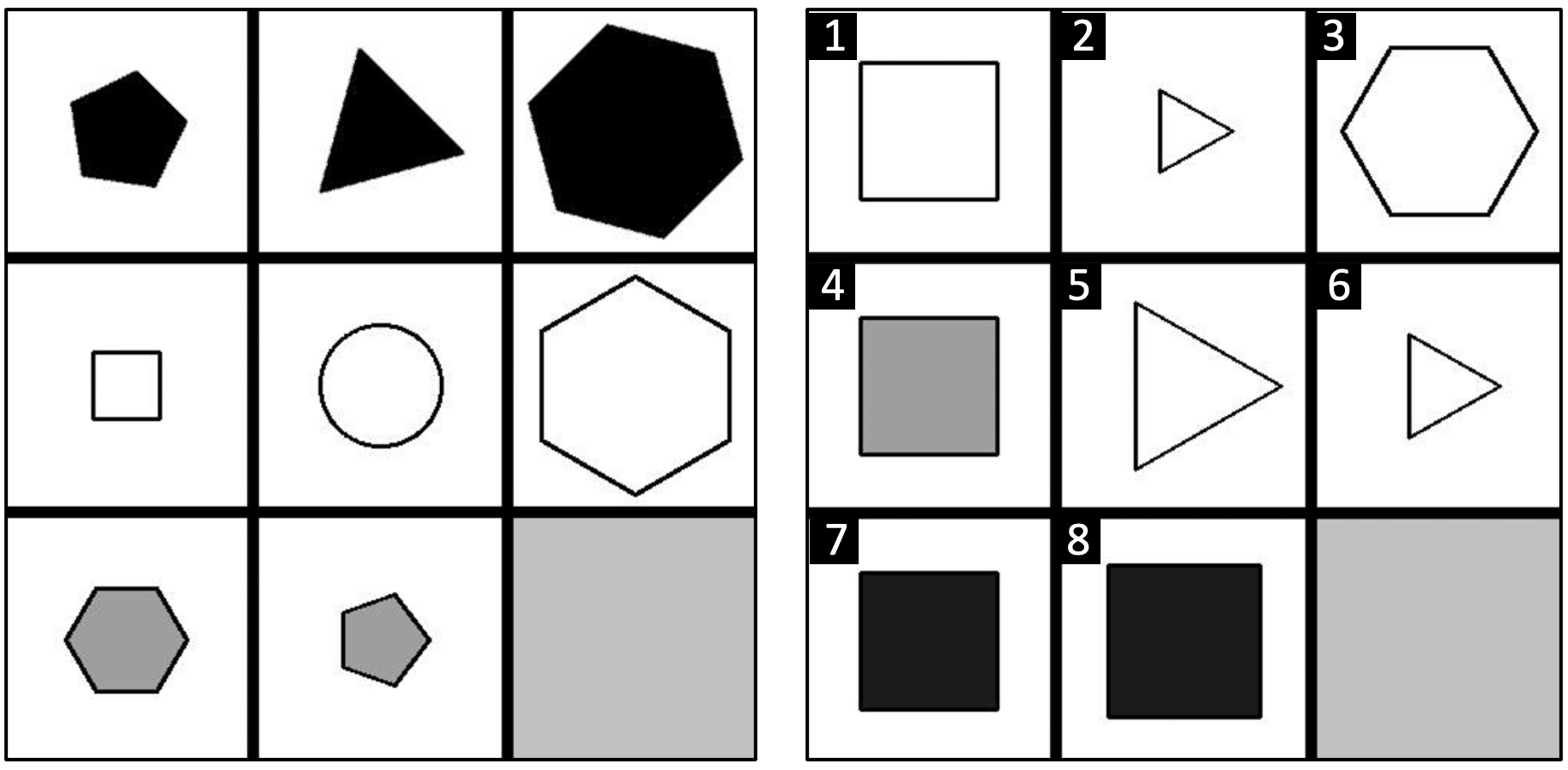} 
    
    \Large \textbf{(b)} 
    
    \vspace*{.1in}

    \includegraphics[width=.55\textwidth]{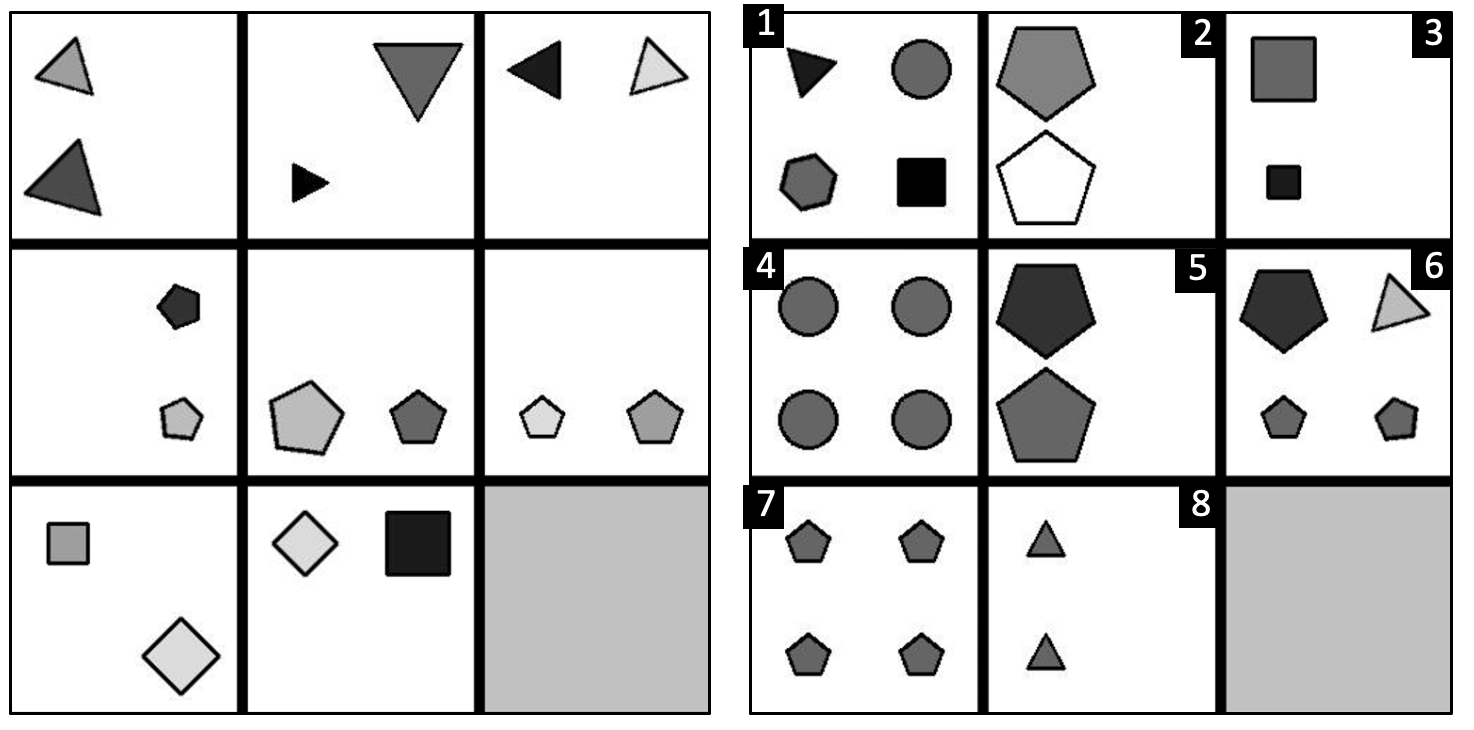} 
    
    \Large \textbf{(c)} 
    
    \vspace*{.1in}
    
    \includegraphics[width=.55\textwidth]{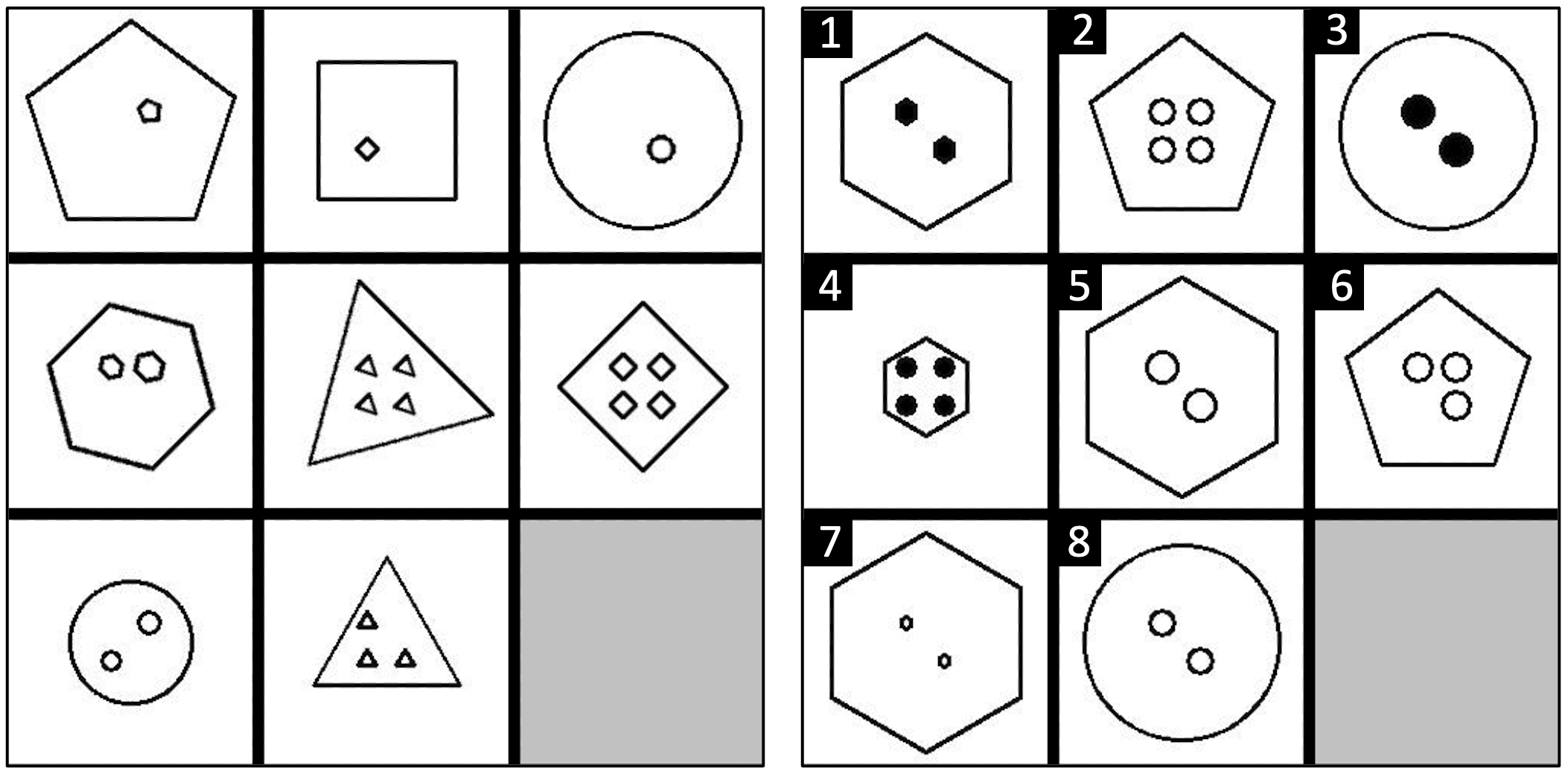} 
    
    \Large \textbf{(d)} 
    
    \caption{Four RAVEN variations on the concept \textit{Sameness}. In Problem (a) all attributes remain constant along each row.  In Problem (b) color stays constant; in Problem (c) number and shape stay constant, and in Problem (d) in each matrix component, the inner object is the same shape as the outer object.  Both SCL and MRNet get the correct answer on (a) and (b), but answer incorrectly on (c) and (d).}
    \label{Sameness}
\end{centering}    

\end{figure*}

\begin{figure*}[h]

\begin{centering}
    \includegraphics[width=.55\textwidth]{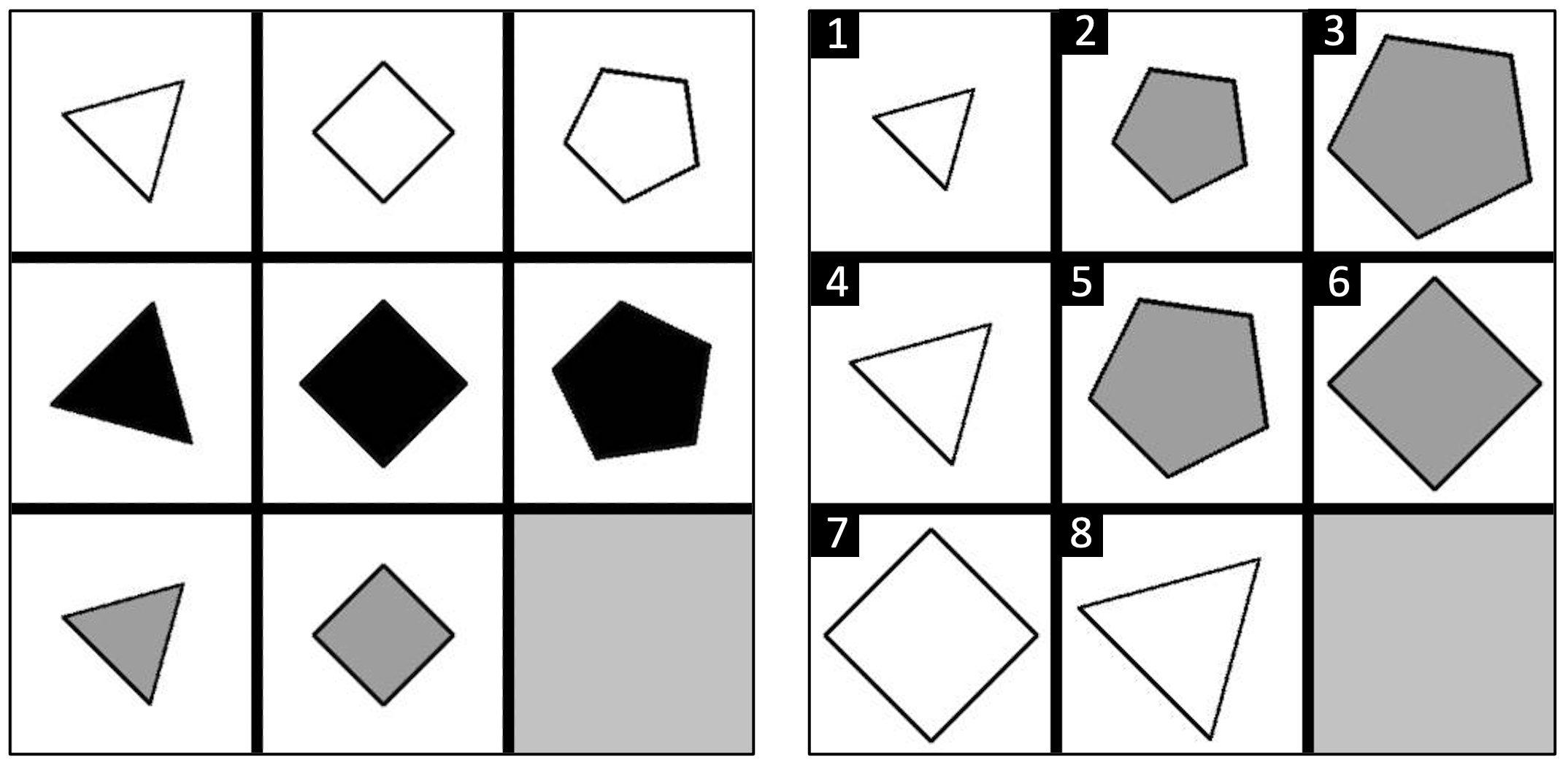} 
    
    \Large \textbf{(a)}
    
    \vspace*{.1in}
    
    \includegraphics[width=.55\textwidth]{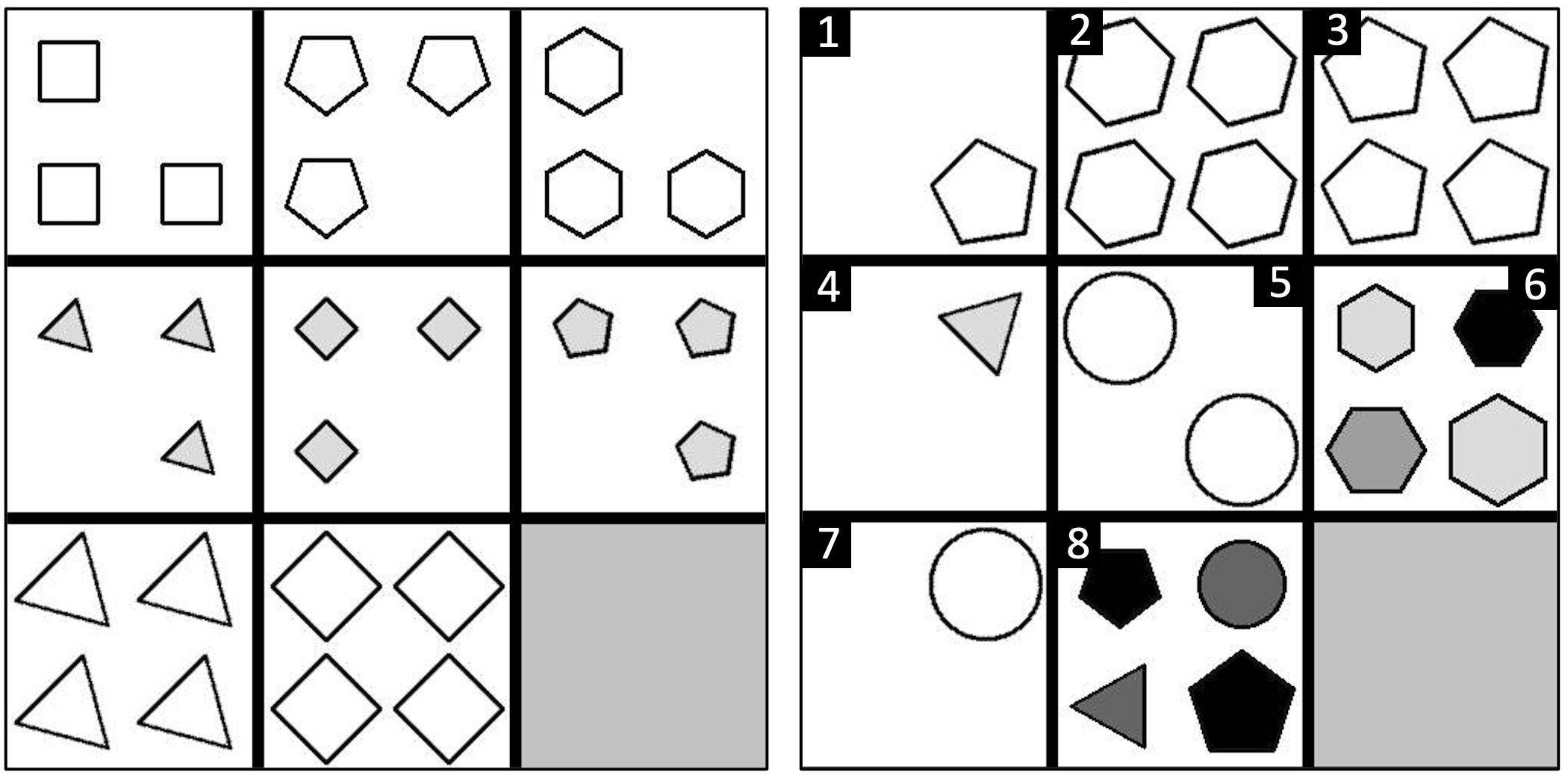} 
    
    \Large \textbf{(b)} 
    
    \vspace*{.1in}

    \includegraphics[width=.55\textwidth]{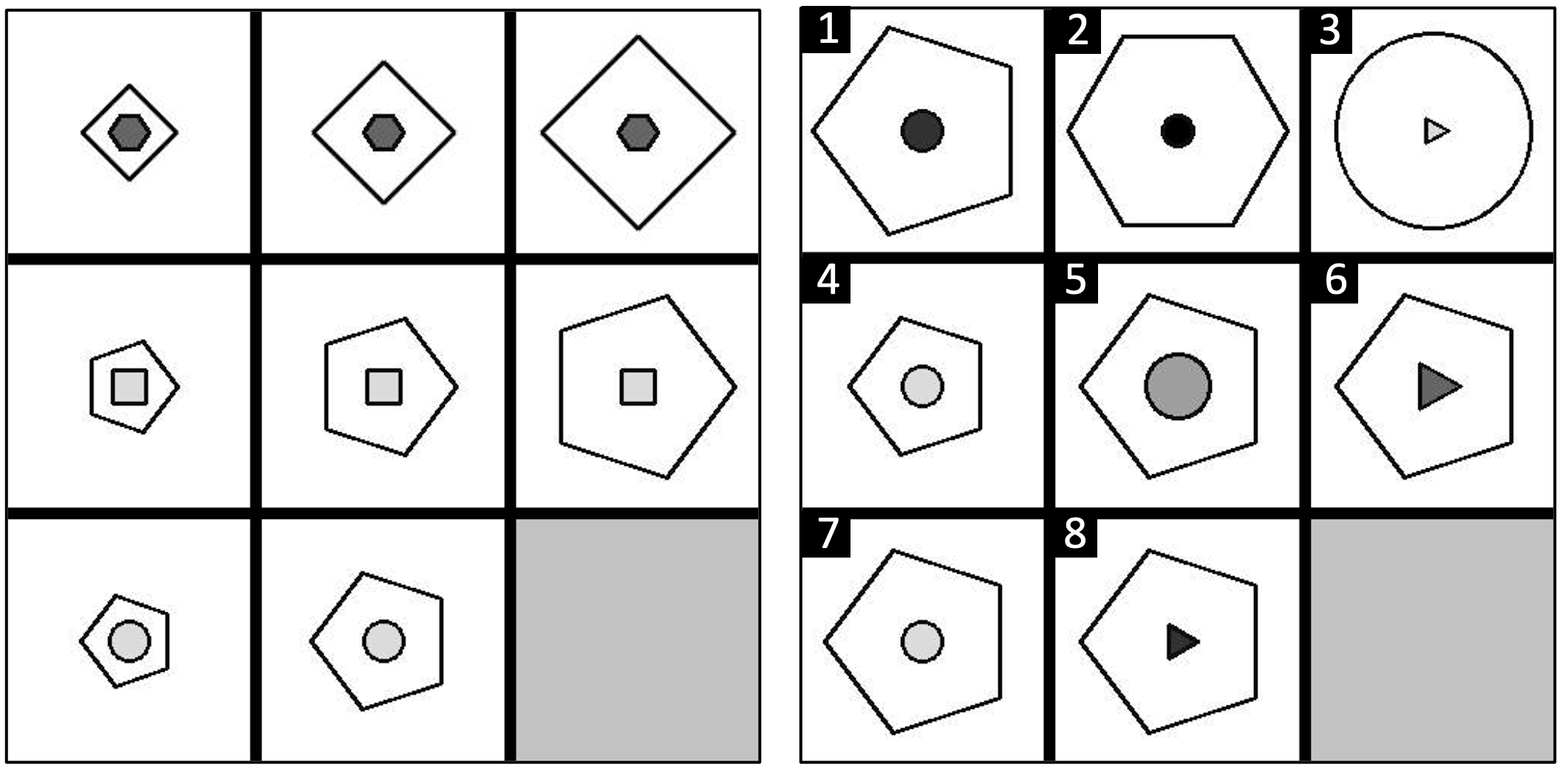} 
    
    \Large \textbf{(c)} 
    
    \vspace*{.1in}
    
    \includegraphics[width=.55\textwidth]{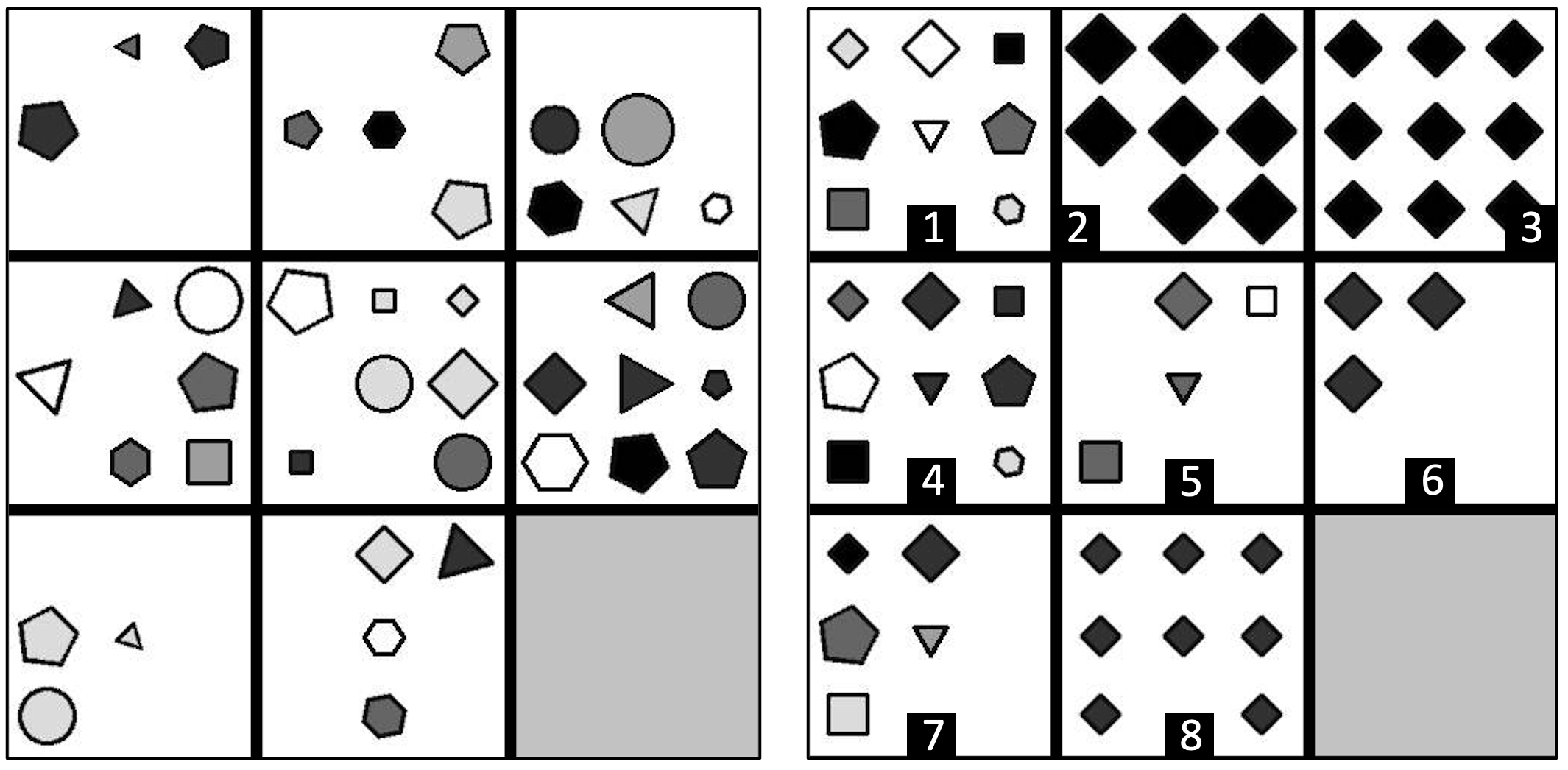} 
    
    \Large \textbf{(d)} 
    
    \caption{Four RAVEN variations on the concept \textit{Progression}. Problem (a) has a progression in the number of sides of the figure along each row; other attributes stay constant.  Problem (b) has the same relationship, but with multiple objects in different positions.  In Problem (c) the progression relation is in the size of the outer figure, and in Problem (d) it is in the number of objects.  SCL chose the correct answer in all but (d) whereas MRNet was correct on (a) and (c) but not on (b) and (d).}
     \label{Progression}
\end{centering}    

\end{figure*}

We first selected two high-performing models from the RAVEN literature: the Multi-scale Relation Network (MRNet, \cite{benny2021scale}) and the Scattering Compositional Learner (SCL, \cite{wu2020scattering}).  For both these systems, the authors made the code publicly available.  We then trained both systems on 30,000 RAVEN training examples---ones that used five of the seven layouts available (Center, 2$\times$2Grid, 3$\times$3Grid, Out-InCenter, and Out-InGrid)\footnote{Because these two models scored each answer individually without any comparison between answers, the models were not affected by the answer-generation bias of the original RAVEN dataset we described above.  Thus we used the original version to train and evaluate them.}  We then evaluated the trained system on 10,000 RAVEN test examples that used these layouts.\footnote{For the sake of time   and simplicity, we omitted the Left-Right and Up-Down layouts, which split each matrix component into two.}  The resulting accuracies on these test examples were 73\% for MRNet and 89\% for SCL.

We then chose two concepts that are present in RAVEN problems: \textit{Sameness} and \textit{Progression}.  Both MRNet and SCL were trained on problems involving some version of these concepts, and both were correct on some instances of these concepts in the RAVEN test set.  In order to probe the degree to which these systems grasp these two concepts, we manually created new problems that systematically vary these concepts, by instantiating these concepts using different attributes.

In all \textit{Sameness} problems, the relevant relationship in each row is that one or more attributes remain constant.  In the RAVEN domain, the possible attributes include shape, size, color (i.e., gray scale), position, row, column, number, angle, and whether one object is inside or outside another object.  Figure~\ref{Sameness} shows four sample Sameness problems from our evaluation set.

In all \textit{Progression} problems, the relevant relationship in each row is an increase (or decrease) in the value of one or more attributes.  Figure~\ref{Progression} shows four sample Progression problems from our evaluation set.  

These samples give a flavor of the problem variations we created around each concept.  Our evaluation consisted of 210 Sameness and 80 Progression problems, designed to instantiate the concepts in ways that we believe would be relatively easy for humans to understand.\footnote{Our Sameness and Progression problems can be downloaded from \url{https://melaniemitchell.me/EBeM2022/RavenVariations.zip}.}  The evaluation results are given in Table~\ref{Raven-Evaluation}.  For both MRNet and SCL, the accuracy on our concept variations are substantially lower than the programs' RAVEN test set accuracy would predict, indicating that their grasp of these general abstract concepts is lacking.  

\begin{table*}
\begin{tabular}{ |c|c|c|c| } \hline
\textbf{Model} & \textbf{RAVEN Test Set (10,000 problems)} & \textbf{Sameness Variations (210 problems)} & \textbf{Progression Variations (80 problems)} \\ \hline 
MRNet &  73\% & 49\% & 44\% \\ \hline    
SCL   &  89\% & 62\% & 68\% \\ \hline 
\end{tabular}
\caption{Accuracy of MRNet and SCL on original RAVEN test set, and on our concept variations.}
\label{Raven-Evaluation}
\end{table*}

\section{Prior Results on ARC}

Deep learning systems such as MRNet and SCL typically lack transparency. Given their large numbers of parameters and training on large IID datasets, they are susceptible to shortcut learning---that is, learning subtle statistical correlations between their input and the correct answers that don't require actual concept understanding \cite{geirhos2020shortcut}.  Such shortcuts are more likely when a system solving problems is allowed to choose from a set of candidate answers, rather than having to generate its own answer.  Moreover, the procedural generation of examples---essential for creating sufficiently large training sets---can be susceptible to overt and subtle biases.

Chollet's ARC dataset \cite{chollet2019intelligence} was created to avoid these pitfalls of deep learning approaches and to be a better method of assessing true abstraction abilities.  Unlike RAVEN and related abstraction datasets, ARC focuses on few-shot learning.  As shown in Figure~\ref{ARC1}, each ARC task can be considered a few-shot-learning task: given a small number of demonstrations, the solver needs to figure out the relevant concept and apply it to the test input grid.  In particular, the solver must \textit{generate} the answer rather than choose from given candidate answers. Moreover, rather than relying on procedurally generated problems, Chollet hand-designed 1,000 tasks, which were used for a competition on the Kaggle website \cite{Chollet-Kaggle}.  Four hundred of the tasks were assigned to a ``training set,'' whose purpose is to give the solver a general idea of what kinds of concepts can be used.  Four hundred additional tasks were assigned to an evaluation set for solvers to assess their abilities, and the 200 hundred remaining tasks make up a unreleased (hidden) test set.  The tasks were carefully designed to capture ``core knowledge'' \cite{spelke2007core} and to assess it in a few-shot, generative framework. 

The Kaggle ARC competition allowed each competing program to generate three answers for each task.  If one of the answers is correct, the program gets credit for solving that task.  Using this metric, the top scorer in the competition was correct on about 21\% of the hidden test cases; the second-place scorer was correct on about 19\%.

\section{Concept-Based Evaluations For ARC}

As a second illustration of our concept-based evaluation approach, we created new ARC tasks to evaluate the Kaggle competition's second-place winner \cite{bleier2020finishing} (whose code was made publicly available).  Here we will call this program \textit{ARC-Kaggle2}.  To probe this program's understanding of concepts in the ARC domain, we selected a number of ARC training tasks that it answered correctly, and identified the concepts a human might have used to solve them. 

\begin{figure*}[t]
\begin{centering}  
  \includegraphics[width=.55\textwidth]{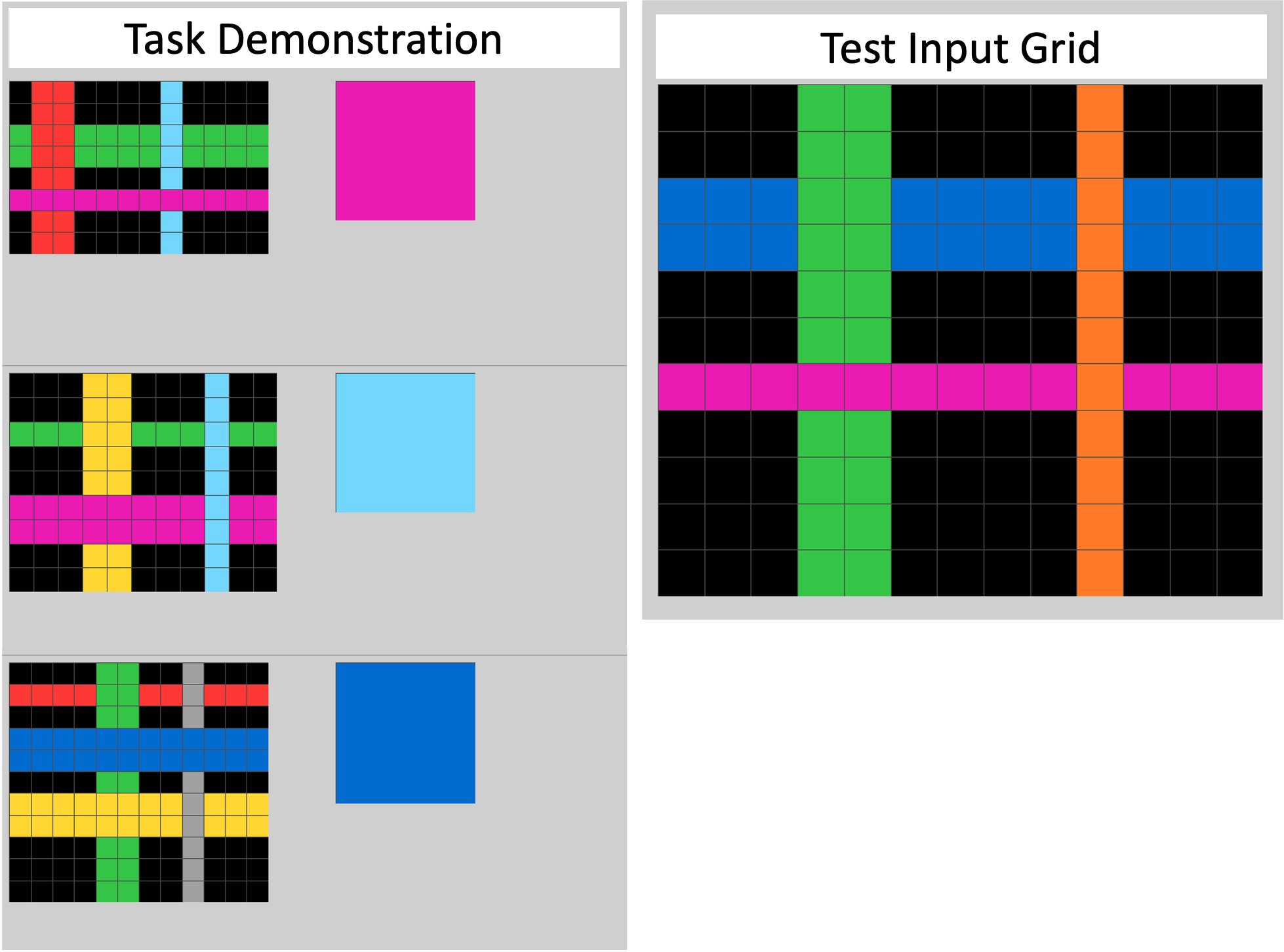}
  
    \Large \textbf{(a)}
    
    \vspace*{.1in}

   \includegraphics[width=.55\textwidth]{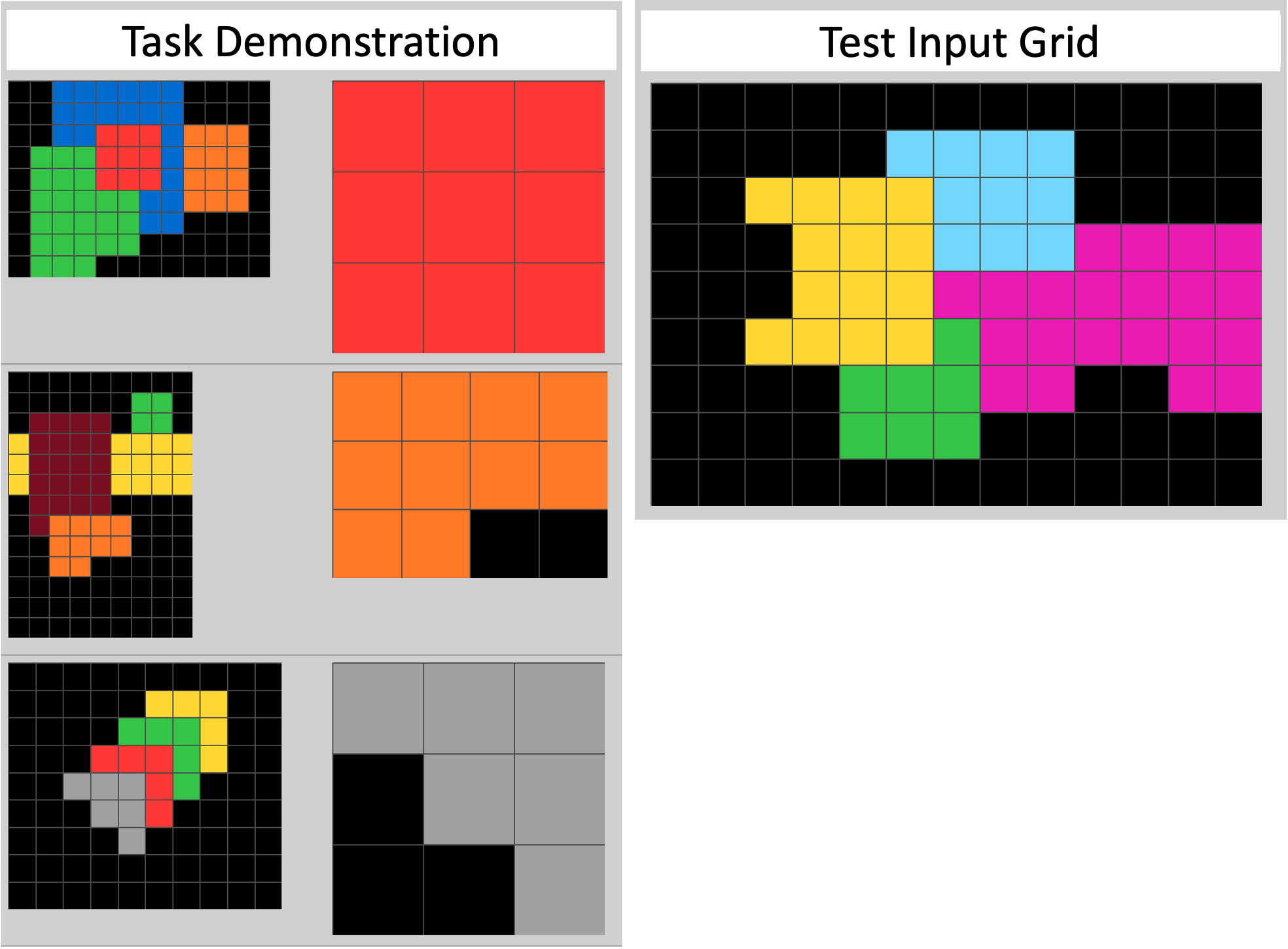}

    \Large \textbf{(b)}
    
    \vspace*{.1in}

    \includegraphics[width=.55\textwidth]{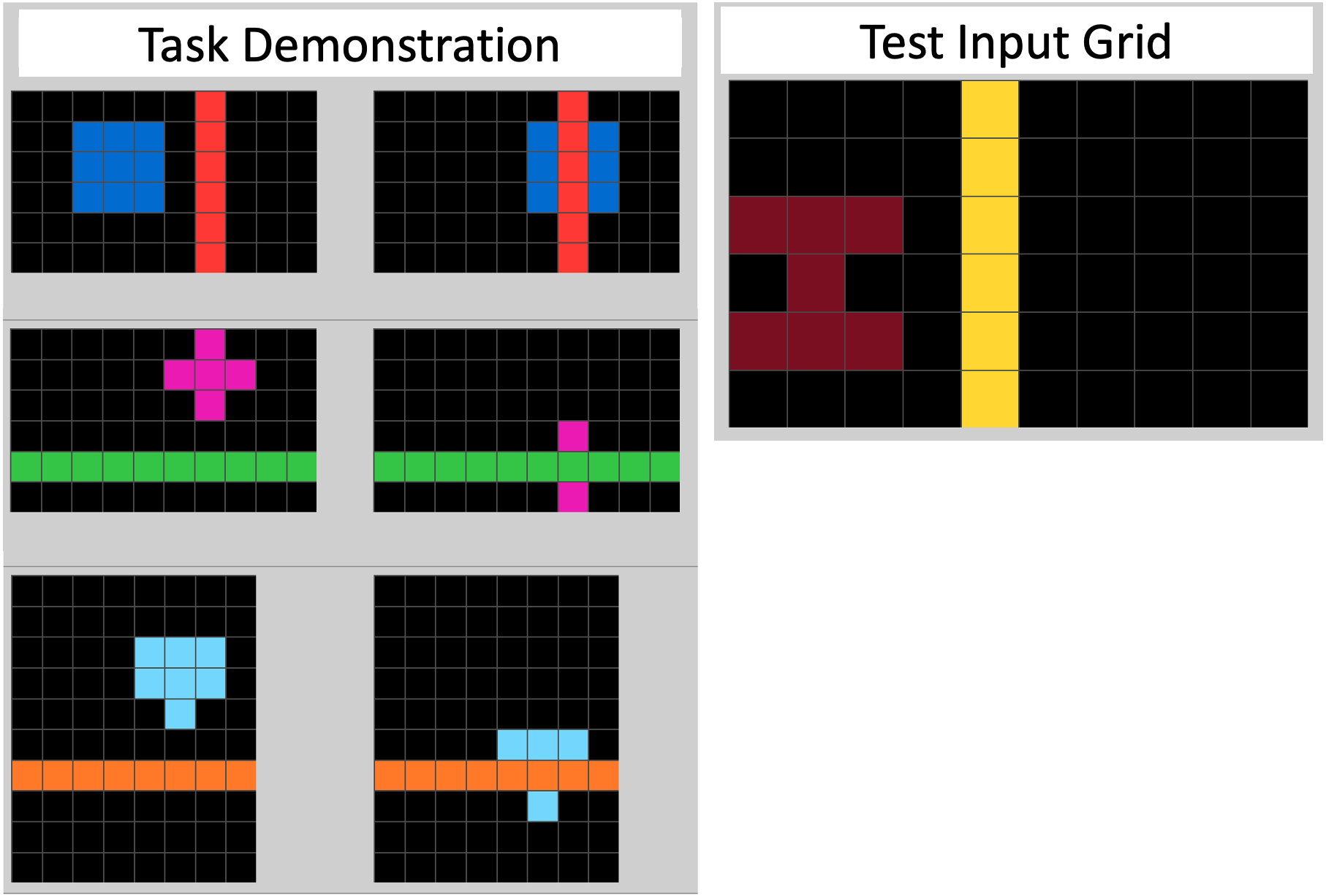}

     \Large \textbf{(c)}
    
    \vspace*{.1in}
    
\end{centering}

  \caption{(a) An ARC task (from the original evaluation set) related to the concept of ``top'' and ``bottom''(or ``above'' and ``below''). The transformation rule is something like ``extract the color of the topmost stripe.'' (b) A sample variation on the ``top/bottom'' concept.  The transformation rule is something like ``extract the the topmost object.''  (c) Another sample variation on the ``top/bottom'' concept.  The transformation rule is something like ``move the object to below the stripe.'' Best viewed in color.}

  \label{ARC-Top-Variations}
  
\end{figure*}

\begin{figure*}[t]

\begin{centering}
    \includegraphics[width=.55\textwidth]{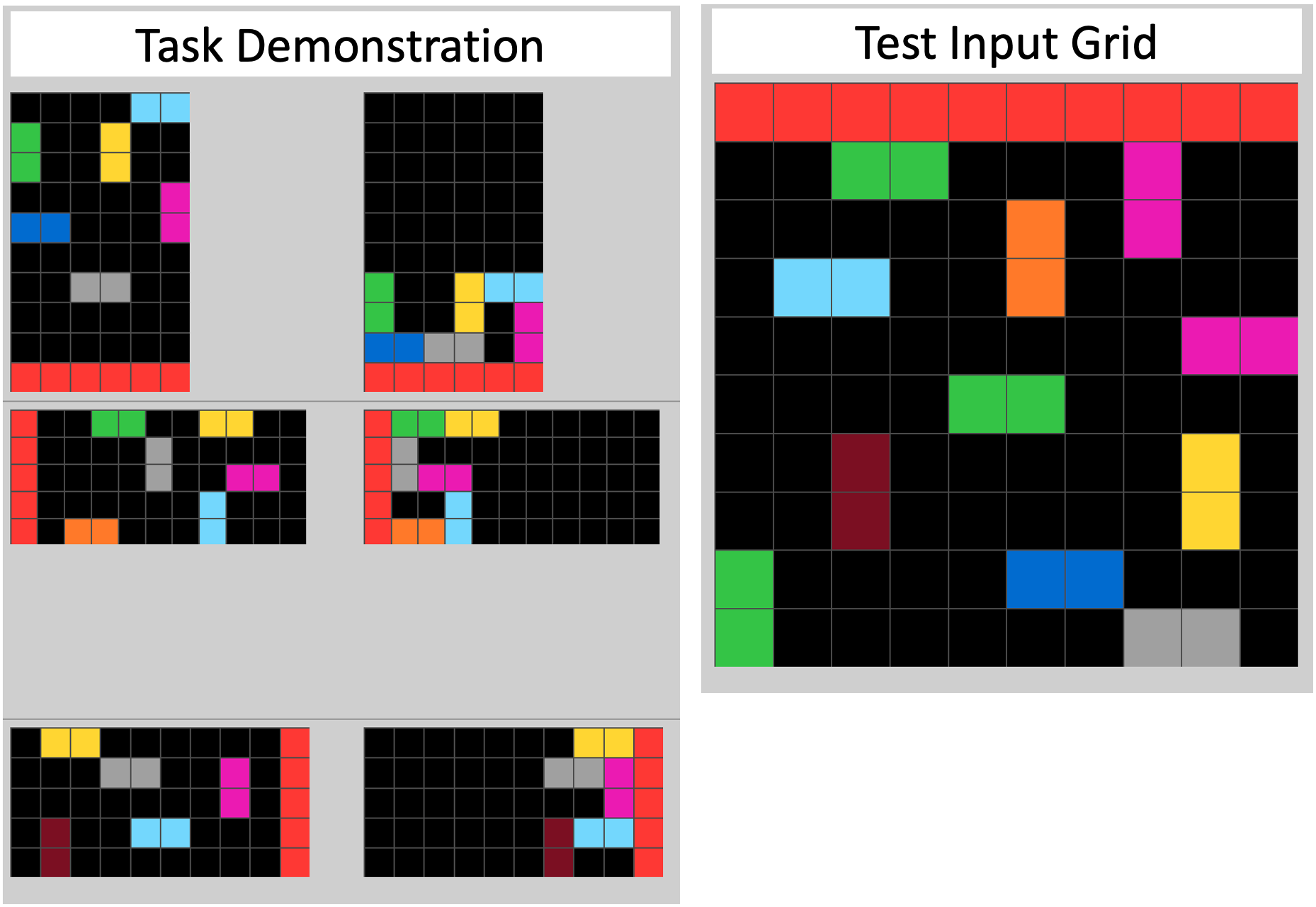}
    
    \Large \textbf{(a)}
    
    \vspace*{.1in}
    
    \includegraphics[width=.55\textwidth]{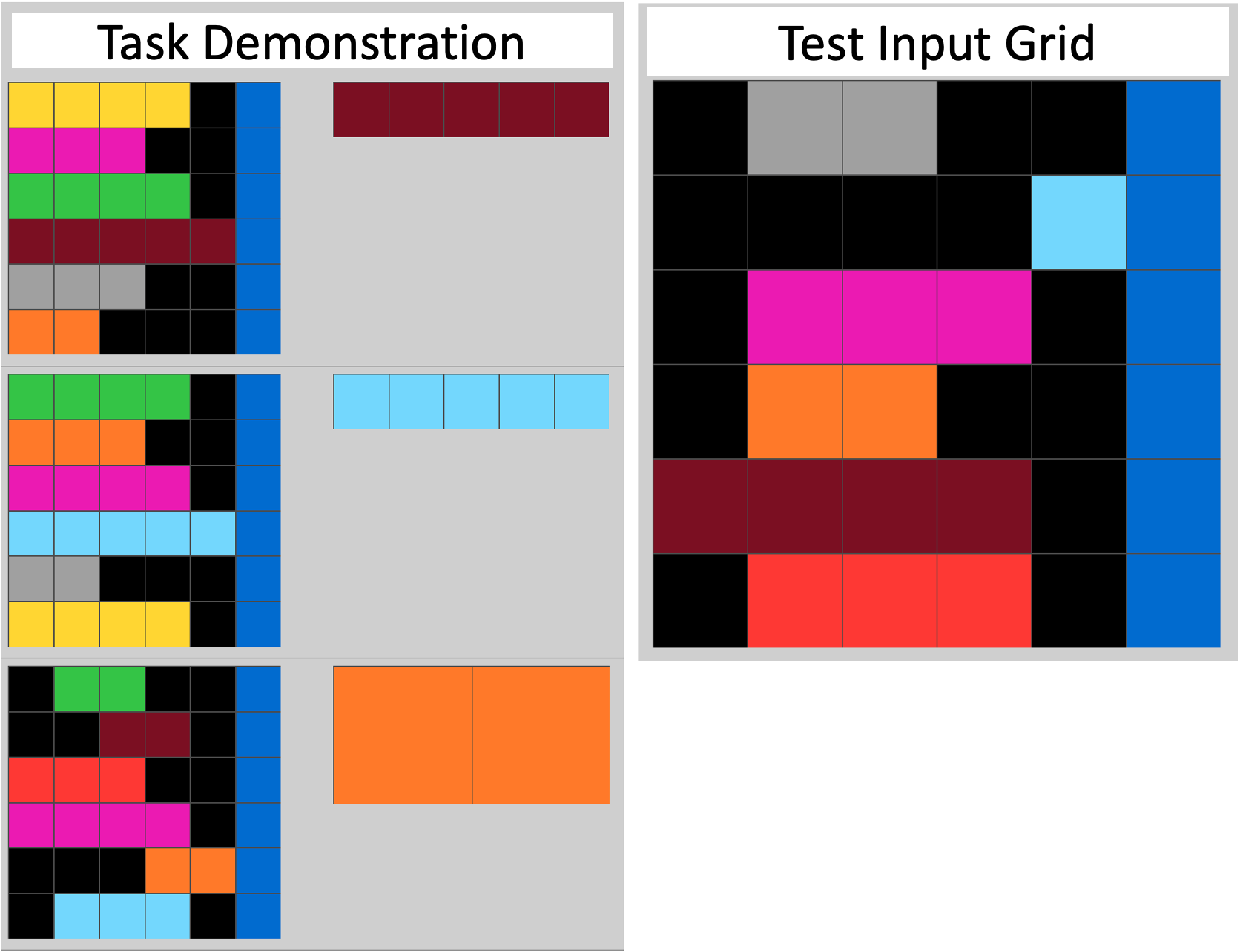} 
    
    \Large \textbf{(b)}
    
    \vspace*{.1in}
    
    \includegraphics[width=.55\textwidth]{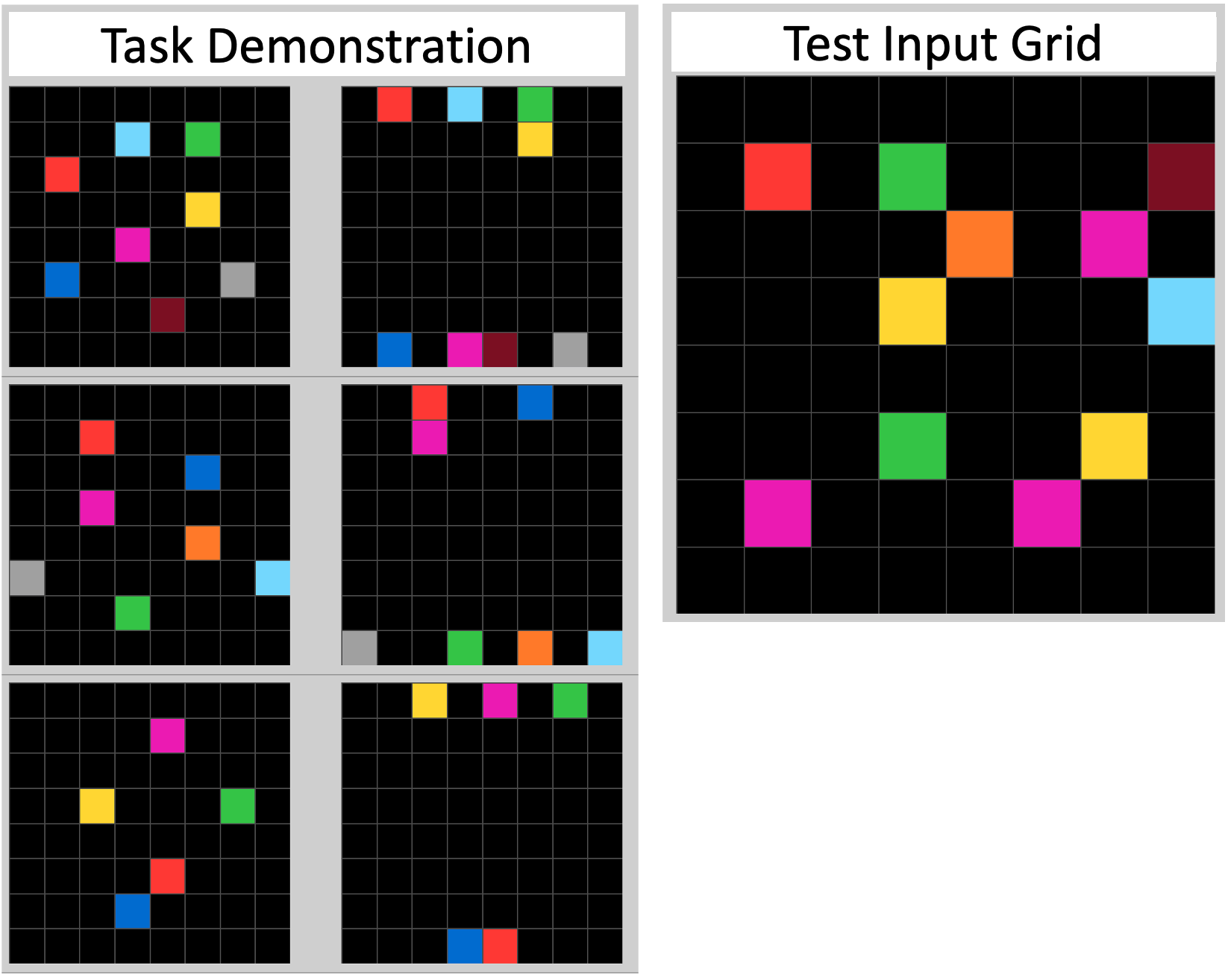}
    
     \Large \textbf{(c)}
    
\end{centering}    
    
    \caption{(a) An ARC task (from the original evaluation set) related to the concept of ``boundary.'' The transformation rule is something like ``Move all objects to the red boundary.'' (b) A sample variation on the ``boundary'' concept.  The transformation rule is something like ``Extract the horizontal stripe that reaches the vertical blue boundary. (c)  Best viewed in color. Another sample variation on the ``boundary'' concept.  The transformation rule is something like ``Move all objects to their closest outer vertical boundary.} 
    \label{ARC-Boundary-Variations}
\end{figure*}

Here we focus on two concepts that appear in the original ARC evaluation set.  The first concept involves spatial notions of ``top'' and ``bottom'' (or ``above'' and ``below'').  The second concept involves the notion of ``boundary.''  Figure~\ref{ARC-Top-Variations}(a) shows a task from the original ARC evaluation set that focuses on the ``top/bottom'' concept: The transformation rule is something like ``Select the color of the topmost stripe.''  ARC-Kaggle2 answered this task correctly.  Figure~\ref{ARC-Boundary-Variations}(a) shows a task from the original ARC evaluation set that focuses on the ``boundary'' concept: The transformation rule is something like ``Move all objects to the red boundary.''  ARC-Kaggle2 also answered this task correctly.

To probe ARC-Kaggle2's grasp of these two concepts, we created variations on ``top/bottom'' and 12 variations on ``boundary.''  To give a flavor of these variations, Figures~\ref{ARC-Top-Variations}(b) and (c) show two of our variants on the ``top/bottom'' concept, and Figures~\ref{ARC-Boundary-Variations}(b) and (c) show two of our variations on the ``boundary'' concept. \footnote{Our ARC task variations can be downloaded from \url{https://melaniemitchell.me/EBeM2022/ARCVariations.zip}.} Table~\ref{ARC-Evaluation} gives the accuracy (given three guesses per task) of ARC-Kaggle2 on our concept variations.  It can be seen that while the program's accuracy on the original ARC test set was 19\%, it appears somewhat better on the ``top/bottom'' concept at 29\% correct, and significantly worse on the ``boundary'' concept at 8\% correct. Given the small number of variations we evaluated the system on, we give these results only as an illustration of our concept-evaluation method; a more thorough evaluation would require many more variations.  

\begin{table*}
  \begin{tabular}{ |c|c|c|c| } \hline
  \textbf{Model} & \textbf{Original ARC Test Set} & \textbf{Top/Bottom Variations (14 tasks)} & \textbf{Boundary Variations (12 tasks)} \\ \hline 
ARC-Kaggle2 &  19\% & 29\% & 8\% \\ \hline 
  \end{tabular}
  \caption{ARC-Kaggle2's accuracy on the original ARC test set as well as on our variations on two concepts.}
  \label{ARC-Evaluation}
\end{table*}

\section{Conclusions and Future Work}

We have argued for assessing AI abstraction programs using systematic concept-based evaluations rather than random training/test splits or IID test sets.  We demonstrated our proposed concept-based evaluation method on existing programs designed to solve problems in the RAVEN and ARC datasets.  Our results indicate that evaluation based on accuracy IID tests set can be uninformative in predicting more generalized performance for a given concept.  In particular, even for concepts present in problems on which the system did well, its performance on concept variations---meant to probe the system's degree of conceptual \textit{understanding}---can be poor.

The results in this paper are meant as an illustration of the method rather than a thorough evaluation; a more complete evaluation would require assessing the systems on many additional concepts, each explored via numerous problem variations. In the future we plan to develop more thorough concept-based evaluation problem suites in not only the RAVEN and ARC domains but in other idealized abstraction and analogy domains for AI systems (e.g., Bongard problems \cite{Bongard1970} and letter-string analogies \cite{Hofstadter1994}).  We also plan to perform human benchmarking studies on these evaluation suites so we can compare human performance with that of machines.

\begin{acknowledgments}
This material is based upon work supported by the National Science Foundation under Grant No.\ 2139983.  Any opinions, findings, and conclusions or recommendations expressed in this material are those of the author and do not necessarily reflect the views of the National Science Foundation. This work was also supported by the Santa Fe Institute.
\end{acknowledgments}

\bibliography{EBeM}

\end{document}